\documentclass[sigconf,natbib=true,anonymous=false]{acmart}

\usepackage{algorithm}
\usepackage{algorithmic}
\usepackage{url}
\usepackage{bbm}
\usepackage{bm}
\usepackage{subcaption}
\usepackage[export]{adjustbox}
\usepackage{amsmath}
\usepackage{mathtools}
\usepackage{multirow}
\usepackage{makecell}
\usepackage{balance}
\AtBeginDocument{%
  \providecommand\BibTeX{{%
    \normalfont B\kern-0.5em{\scshape i\kern-0.25em b}\kern-0.8em\TeX}}}

\copyrightyear{2024}
\acmYear{2024}
\setcopyright{acmlicensed}\acmConference[WWW '24 Companion]{Companion Proceedings of the ACM Web Conference 2024}{May 13--17, 2024}{Singapore, Singapore}
\acmBooktitle{Companion Proceedings of the ACM Web Conference 2024 (WWW '24 Companion), May 13--17, 2024, Singapore, Singapore}
\acmDOI{10.1145/3589335.3651927}
\acmISBN{979-8-4007-0172-6/24/05}
\begin{document}
\settopmatter{authorsperrow=4}

\title{Prompt Optimizer of Text-to-Image Diffusion Models for Abstract Concept Understanding}


\author{Zezhong Fan}
\authornote{Both authors contributed equally to this research.}
\affiliation{%
  \institution{Walmart Global Tech}
  \city{Sunnyvale}
  \state{California}
  \country{USA}}
\email{zezhong.fan@walmart.com}

\author{Xiaohan Li}
\authornotemark[1]
\affiliation{%
  \institution{Walmart Global Tech}
  \city{Sunnyvale}
  \state{California}
  \country{USA}}
\email{xiaohan.li@walmart.com}

\author{Chenhao Fang}
\authornote{Work done while at Walmart.}
\affiliation{%
  \institution{University of Wisconsin-Madison}
  \state{Wisconsin}
  \country{USA}
}
\email{chenhao.fang@outlook.com}

\author{Topojoy Biswas}
\affiliation{%
  \institution{Walmart Global Tech}
  \city{Sunnyvale}
  \state{California}
  \country{USA}}
\email{topojoy.biswas@walmart.com}

\author{Kaushiki Nag}
\affiliation{%
  \institution{Walmart Global Tech}
  \city{Sunnyvale}
  \state{California}
  \country{USA}}
\email{kaushiki.nag@walmart.com}

\author{Jianpeng Xu}
\affiliation{%
  \institution{Walmart Global Tech}
  \city{Sunnyvale}
  \state{California}
  \country{USA}}
\email{jianpeng.xu@walmart.com}

\author{Kannan Achan}
\affiliation{%
  \institution{Walmart Global Tech}
  \city{Sunnyvale}
  \state{California}
  \country{USA}}
\email{kannan.achan@walmart.com}

\renewcommand{\shortauthors}{Zezhong Fan et al.}

\begin{abstract}
    The rapid evolution of text-to-image diffusion models has opened the door of generative AI, enabling the translation of textual descriptions into visually compelling images with remarkable quality. However, a persistent challenge within this domain is the optimization of prompts to effectively convey abstract concepts into concrete objects. For example, text encoders can hardly express "peace", while can easily illustrate olive branches and white doves. This paper introduces a novel approach named Prompt Optimizer for Abstract Concepts (POAC) specifically designed to enhance the performance of text-to-image diffusion models in interpreting and generating images from abstract concepts. We propose a Prompt Language Model (PLM), which is initialized from a pre-trained language model, and then fine-tuned with a curated dataset of abstract concept prompts. The dataset is created with GPT-4 to extend the abstract concept to a scene and concrete objects. Our framework employs a Reinforcement Learning (RL)-based optimization strategy, focusing on the alignment between the generated images by a stable diffusion model and optimized prompts. Through extensive experiments, we demonstrate that our proposed POAC significantly improves the accuracy and aesthetic quality of generated images, particularly in description of abstract concepts and alignment with optimized prompts. We also present a comprehensive analysis of our model's performance across diffusion models under different settings, showcasing its versatility and effectiveness in enhancing abstract concept representation.

\end{abstract}

\ccsdesc[500]{Computing methodologies~Computer vision}
\ccsdesc[500]{Computing methodologies~Artificial intelligence}

\keywords{Image generation, Diffusion models, Prompt optimization, Abstract concepts}

\maketitle

\section{Introduction}
\begin{figure}
    \centering
    \includegraphics[width=0.45\textwidth]{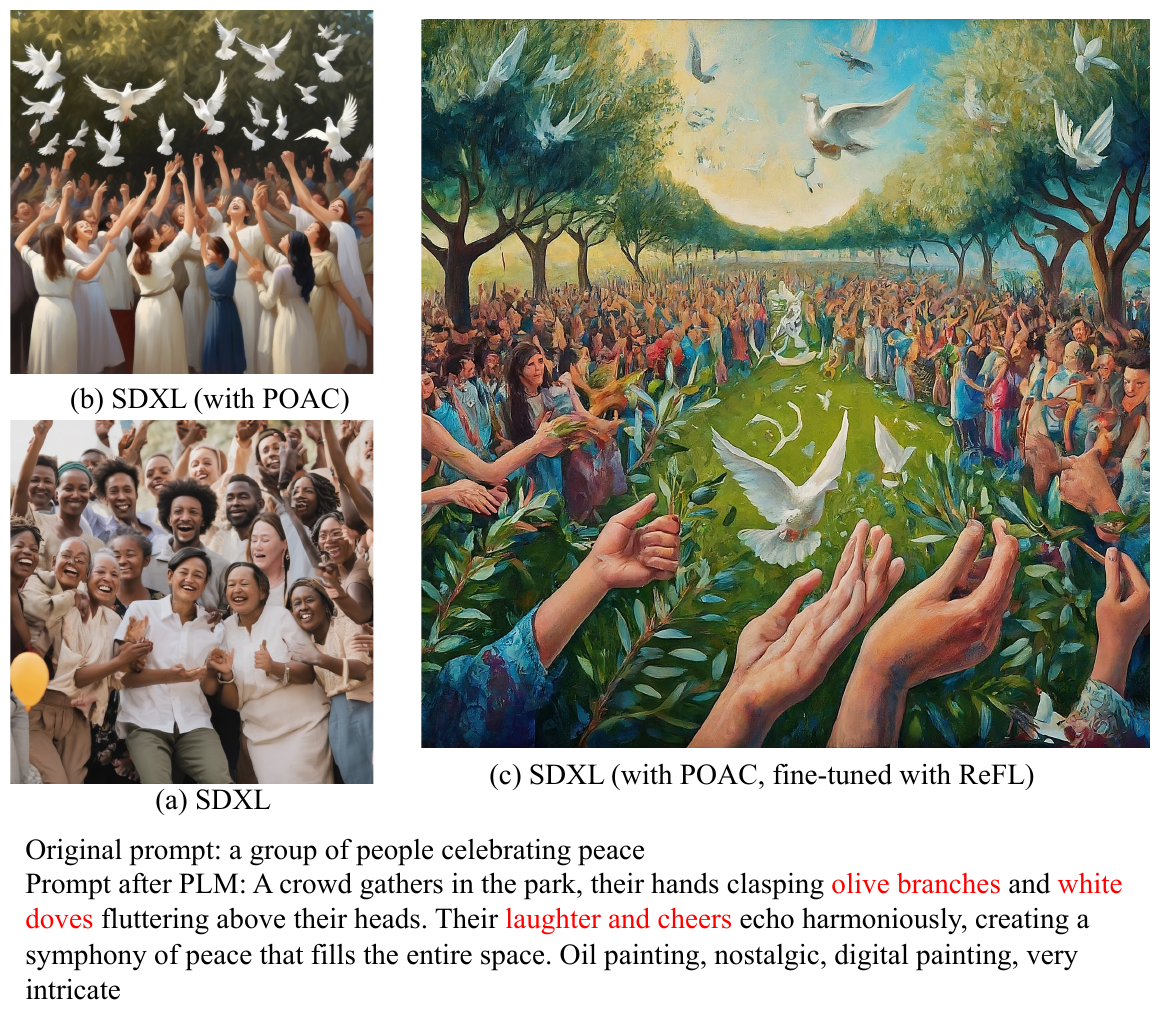}
    \caption{Examples of image generation with or without prompt optimizer. (a) SDXL: only input the original prompt. (b) SDXL (with POAC): input the prompt optimized by POAC to SDXL. (c) Generate the image with the fine-tuned SDXL with ReFL and optimized prompt.}
    \label{fig:peace}
\end{figure}

In human daily communication, concepts serve as a medium that conveys diverse ideas. People usually utilize two different ways to express their thoughts: concrete objects and abstract concepts. Concrete objects refer to tangible, observable, and measurable phenomena or entities and they are based on sensory perception and can be directly experienced like a tree or rain. On the other hand, abstract concepts refer to ideas or theories that are not based on physical or concrete phenomena, such as "peace" and "courage". They are intangible and cannot be directly observed or measured. Abstract concepts play an important role in expressing complex emotions and shaping ethical and moral values. In real life, to alleviate the obstacles caused by abstract concepts in daily communication, people often explain them with their related concrete objects and visible objects. For example, as shown in Figure~\ref{fig:peace} when people introduce the "Peace" concept, they often refer to olive branches and white doves to help others understand. 

Recent cutting-edge Text-to-Image (T2I) generation models including Stable Diffusion~\cite{rombach2022high} ,DALL•E~\cite{ramesh2021zero}, Imagen~\cite{saharia2022photorealistic} and Dreambooth~\cite{ruiz2023dreambooth} have achieved outstanding achievements in generating high-resolution realistic images. However, these models are trained with large-scale text-image pairs datasets which mainly concentrate on concrete and physical objects and attributes, thus failing to generate accurate visualizations or descriptions for abstract ideas as their narrow understanding of these abstract concepts. 



There are three main challenges in comprehending and explaining the abstract concept in images using T2I generation models. First, the current research has primarily examined subject-driven T2I models~\cite{ruiz2023dreambooth, kumari2023ablating, ma2023subject, kumari2023multi} that concentrate solely on concrete objects in image generation. However, expressing abstract concepts in images is challenging as they require association with concrete objects. Currently, there is limited research on the translation of abstract concepts to concrete objects. Second, the most recent T2I model for abstract concepts~\cite{liao2023text} enhances prompt optimization by utilizing a Large Language Model (LLM)~\cite{brown2020language} and expanding abstract concepts to encompass multiple concrete objects. However, this optimization solely focuses on the language aspect and does not align with the image generation process. Third, the utilization of LLMs requires the use of a large GPU server or purchasing services from a third-party company, both of which are costly and not easily scalable. In this paper, we intend to develop a compact language model that specifically focuses on prompt optimization to enhance scalability.


To address the above challenges, we propose a novel \textbf{P}rompt \textbf{O}ptimization framework of Text-to-Image Generation for \textbf{A}bstract \textbf{C}oncepts (\textbf{POAC}) that purposes on bridging the diffusion model and the abstract concepts while ensuring the generation faithfulness and maintaining aesthetic appeal.  Inspired by the recent advances in prompt optimization and abstract concept understanding~\cite{liao2023text, hao2022optimizing, lee2024optimizing}, we first trained a Prompt Language Model (PLM) with Supervised Fine-Tuning (SFT) based on a pre-trained language model such as GPT2~\cite{radford2019language} on a collection of text inputs with abstract concepts and optimized prompt with physical scenes and objects corresponding to them. Then we employ Reward Feedback Learning (ReFL) based on Reinforcement Learning (RL) to align the PLM with the diffusion model. The ReFL is applied to fine-tune the diffusion model using generated images and optimized prompts generated by PLM as well as original prompts with different weights to improve the alignment between human intents and generated images. In Figure~\ref{fig:peace}, we can see that the image (c) generated by fine-tuned SDXL with ReFL has a better alignment with the prompt as it has both white doves and olive branches while image (b) only includes white doves. The reward used in both stages is a combination of relevance scores~\cite{radford2021learning} and aesthetic scores~\cite{xu2023imagereward}. To help the model better understand the abstract concept, we also create a dataset with the help of GPT-4~\cite{achiam2023gpt} to map abstract concepts to concrete objects.

We conduct experiments with one of the cutting-edge T2I diffusion models, Stable Diffusion XL (SDXL)~\cite{podell2023sdxl}. We evaluate our methods utilizing the latest benchmark and human evaluation. The experimental results show that POAC can precisely express and understand abstract concepts with visible details while ensuring aesthetic appeal and faithfulness of generated images. Moreover, we also find that two-stage reinforcement learning plays a vital role in improving image quality and aligning with human preference. 

The main contributions of our works are as follows:
\begin{itemize}
    \item With the assistance of GPT-4, a dataset is constructed to correlate abstract concepts with concrete objects.
    \item We fine-tune a Prompt Language Model (PLM), which rewrites the abstract concepts in the original prompts to concrete objects in the optimized prompts.
    \item We also fine-tune the SDXL with ReFL to align the optimized prompts and original prompts with the generated images.
\end{itemize}
    





\section{Related works}
\subsection{Text-to-Image Generation}
Text-to-Image generation is a crucial task which aims to generate images based on user-input textual descriptions in computer vision. The existing models in this domain can be categorized as either autoregressive ~\cite{gafni2022make,yu2022scaling, ramesh2021zero} or diffusion models ~\cite{nichol2021glide, ramesh2022hierarchical, rombach2022high, saharia2022photorealistic}, as evidenced by prior research. Notably, DALL-E ~\cite{ramesh2021zero} stands out as a significant advancement in autoregressive models due to its impressive zero-shot capabilities. However, diffusion models have also shown very promising results. Based on prior works in the diffusion models ~\cite{dhariwal2021diffusion, ho2022cascaded}, GLIDE model ~\cite{nichol2021glide} introduces an innovative approach to conditioning the diffusion model based on an input text caption. Additionally, DALL-E 2 ~\cite{ramesh2022hierarchical} model is further improved by incorporating a supplemental CLIP image embedding, which enhances the model's diversity. Other certain efforts, such as Stable Diffusion ~\cite{rombach2022high}, prioritize computational efficiency by initially representing input images as low-dimensional latent codes. However, due to most of training prompts of these models only including concrete objects in real-life and lack of mapping of abstract concepts and images, it is still a challenge for diffusion models to fully understand abstract concept directly. 

To further align text-to-image generation models with human understandings, many efforts have been made to alleviate the gap between models and human preference ~\cite{hao2022optimizing, xu2023imagereward, wu2023better, dong2023raft, lee2023aligning}. For example, Lee et al.~\cite{lee2023aligning} emphasize text alignment with diffusion models, utilizing a reward model that was trained on datasets annotated by human evaluators to fine-tune the text-to-image model. Similarly, ImageReward ~\cite{xu2023imagereward} provides a comprehensive human preference reward model that align text with images from aesthetics, toxicity, and biases. To enhance the original prompt, Hao et al.~\cite{hao2022optimizing} propose a prompt adaptation framework to train a language model with a reinforcement learning framework. However, there are still a gap of aligning the human preferences with abstract concepts.

\subsection{Concept Representation in Image Generation}
The field of image generation has witnessed a notable surge in interest in the representation of concepts. This includes efforts in concept customization~\cite{ruiz2023dreambooth, gal2022image, kumari2023multi, wei2023elite} which have sought to enrich Text-to-Image (T2I) models with new concepts. These concepts typically focus on incorporating specific entities, ranging from novel creations to personalized items from everyday life, such as a pet dog. Additionally, the area of concept disambiguation, as explored in \cite{mehrabi2022elephant}, addresses the challenges posed by the syntactic ambiguity in human instructions, which can obscure the intended meanings and relationships of physical items, without extensively examining the nuanced differences among abstract concepts. In conclusion, recent research in image generation has focused on the visualization of tangible concepts instead of abstract ones. Our objective is to narrow this research gap by deepening the exploration of abstract concepts in image generation.

\subsection{Prompt Optimization for Diffusion Models}
As large-scale models reliant on textual inputs continue to evolve~\cite{chen2023knowledge, liu2024cliqueparcel, maragheh2023llm, fang2024llm}, the substantial resources required for their training and fine-tuning present a significant barrier for many researchers. In response, prompt optimization has been recognized as an effective strategy. This approach enhances image generation quality without the need for altering the underlying model architecture or engaging in labor-intensive training processes for T2I models. The area of prompt optimization is still in its early stages and suffers from a scarcity of exhaustive studies. Practical advice on writing effective T2I prompts is primarily shared through blogs and manuals~\cite{oppenlaender2022prompt, oppenlaender2022taxonomy, pavlichenko2023best}, which summarize the essential elements and descriptive terms that enhance image style. Prompt optimization can be applied within textual or embedding dimensions, including soft tuning methods~\cite{lester2021power}, with some researchers focusing on developing optimization models that closely integrate with T2I models~\cite{hao2022optimizing}. While existing efforts in T2I prompt optimization~\cite{lee2024optimizing} largely concentrate on refining the stylistic and aesthetic qualities of generated images, our study intends to focus on advancing the visual representation of abstract concepts.
\section{Methodology}

\begin{figure*}
    \centering
    \includegraphics[width=0.8\textwidth]{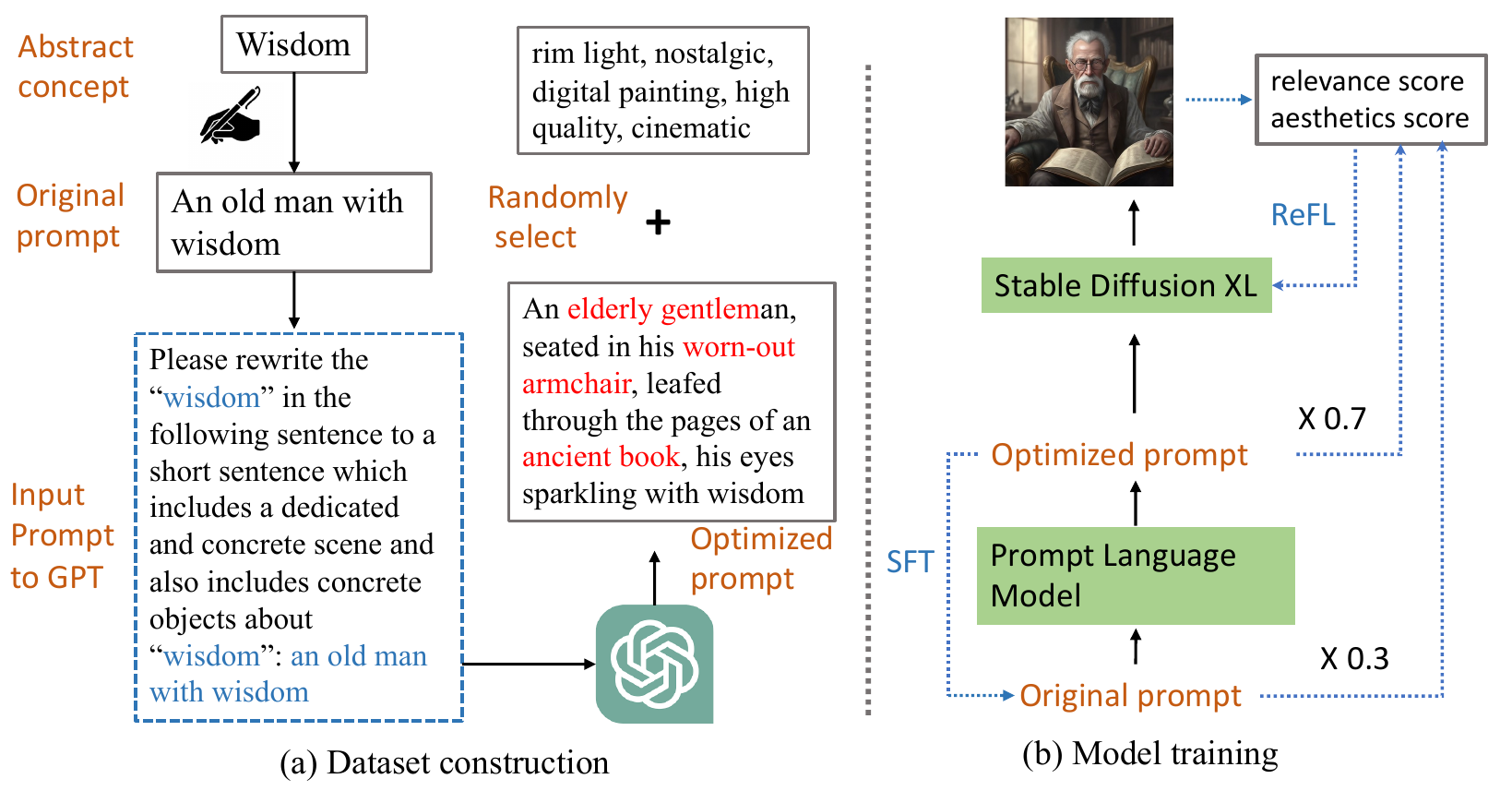}
    \caption{(a) The dataset construction process. We manually rewrite the abstract concept "wisdom" to a short prompt. With the help of GPT, we prompt is optimized with detailed and concrete objects (in red). The art styles are randomly selected and added to the optimized prompt. (b) The training process of Prompt Language Model (PLM) and Stable Diffusion XL (SDXL). PLM is fine-tuned with original and optimized prompts by Supervised Fine-Tune (SFT). The SDXL is fine-tuned by Reward Feedback Learning (ReFL) to align the prompts and the image.}
    \label{fig:model}
\end{figure*}

The aims of  Prompt Optimizer for
Abstract Concepts (POAC) is to automatically enrich abstract concepts with their corresponding concrete objects and feed optimized prompts into a diffusion model to generate aesthetic and faithful images.  Figure 1 presents the overview of our framework. POAC is composed of two parts: 1) Prompt Language Model (PLM) is built based on a pre-trained language model (GPT2)~\cite{radford2019language} and 2) Stable Diffusion XL (SDXL), which is a large-scale diffusion model ~\cite{podell2023sdxl}. We also construct a dataset containing pairs of abstract concepts and their corresponding concrete objects. With Supervised Fine-Tuning(SFT), we fine-tune the PLM with our constructed dataset. Then we conduct reinforcement learning on the SDXL to maximize target reward to improve diffusion model understanding of abstract concepts and ensure alignment with prompts and aesthetics of generated images. 

\subsection{Datasets}
We first collect top-500 abstract concepts from IBM Concept Abstractness\footnote{https://developer.ibm.com/exchanges/data/all/concept-abstractness/} based on their degree of abstractness. Then, we manually extend each abstract concept into a short prompt including it as the source input to PLM. For example, as shown in Figure~\ref{fig:model}, we extend "wisdom" to "an old man with wisdom". Finally, we leverage the strong knowledge and understanding of abstract concepts of GPT-4~\cite{achiam2023gpt} to rewrite the source input into a new prompt with a dedicated scene and concrete objects that can express the abstract concepts clearly and effectively as target outputs. From high-quality human-engineered prompts from various diffusion models in Internet~\cite{oppenlaender2022prompt, pavlichenko2023best}, we can discover that most prompts are composed of two parts: main content that describes the user’s intention, and some modifiers expressing image style such as artist name and art style. To ensure the quality and diversity of generated images, we randomly add modifiers to target outputs.  


\subsection{Supervised Fine-tuning}
We first fine-tune PLM initialized with a pre-trained language model (GPT-2~\cite{radford2019language}) on a set of original and optimized prompt pairs. A parallel prompt dataset $\mathcal{D} = {(\mathbf{x}, \mathbf{y})}$ contains prompt pairs of short prompts with abstract concepts and rewritten prompts with corresponding concrete objects. The training objective is to maximize the log-likelihood:  
\begin{equation}
    \mathcal{L}_{SFT} = -\mathbb{E}_{(\mathbf{x},\mathbf{y})\sim \mathcal{D}}\log p(\mathbf{y}|\mathbf{x})
\end{equation}
where PLM is trained to be used to generate the input prompts of the diffusion model. 

\subsection{Reward Function}
\label{sec:reward}
To ensure the faithfulness and quality of generated images, we design our reward function from two perspectives: relevance to abstract concepts and aesthetics.
The goal motivates us to define the reward function $\mathcal{R}(\cdot)$ from the above two perspectives. First, we use CLIP scores~\cite{radford2021learning} to measure the relevance of between generated images and prompts containing abstract concepts as well as optimized prompts with concrete objects. The relevance score is defined as below: 

\begin{equation}
    \mathcal{R}_{rel}(\mathbf{x}, {\mathbf{y}}) =  \mathbb{E}_{i_{\mathbf{y}}\sim \mathcal{G}(\mathbf{y})}[f_{rel}(\mathbf{x}, \mathbf{y},i_{\mathbf{y}})]
\end{equation}

\begin{equation}
    f_{rel}(\mathbf{x}, i_{\mathbf{y}}) = 0.3 * g_{CLIP}(\mathbf{x}, i_{\mathbf{y}}) + 0.7 *g_{CLIP}(\mathbf{y}, i_{\mathbf{y}})
\end{equation}
where $i_{\mathbf{y}}\sim \mathcal{G}$ represents sampling images $i_{\mathbf{y}}$ from the diffusion model $\mathcal{G}$ with $\mathbf{y}$ as input prompts. $g_{CLIP}$ is the CLIP similarity function. We determine different weights for CLIP scores of original prompts with abstract concepts and optimized prompts with concrete objects to 0.3 and 0.7 because we want to encourage the model to focus more on physical details while ensuring intentions from original abstract concepts. 

Then, to measure the aesthetic preference of images, we use ImageReward~\cite{xu2023imagereward} score which is a T2I human preference reward model. It is trained based on a large human-annotated aesthetic preference dataset containing 137k expert comparisons. The aesthetic score is defined as: 
\begin{equation}
    \mathcal{R}_{aes}(\mathbf{y}) = \mathbb{E}_{i_y\sim \mathcal{G}(\mathbf{y})}[g_{aes}(i_{\mathbf{y}})],
\end{equation}
where $g_{aes}$ denotes ImageReward score.

Finally we define the overall reward by combining the above scores:
\begin{equation}
    \mathcal{R}(\mathbf{x}, \mathbf{y}) = \mathcal{R}_{rel}(\mathbf{x}, \mathbf{y}) + \mathcal{R}_{aes}(\mathbf{y})
\end{equation}

\subsection{Reward Feedback Learning}
To make diffusion models better align with abstract concepts, we adopt a strategy known
as Reward Feedback Learning (ReFL)~\cite{xu2023imagereward}. During the Reward Feedback Learning stage, ReFL is employed to fine-tune diffusion models based on the reward function in 3.3. The reward functions are employed to back-propagate and update the diffusion parameters after a predetermined range of steps. In practice, to mitigate the over-optimization issue and stabilize
the fine-tuning, we re-weight reward loss and regularize with pre-training loss. 
After ReFL, the diffusion model is optimized with a focus on aligning generated images with abstract concepts. The loss of ReFL is formulated as

\begin{equation}
    \mathcal{L}_r = \lambda \mathbb{E}_{y\sim \mathcal{Y}}(\phi(\mathcal{R}(\mathbf{x}, \mathbf{y}))),
\end{equation}
\begin{equation}
    \mathcal{L}_{pre} = \mathbb{E}_{(y, i_y)\sim \mathcal{G}(y)}(\mathbb{E}_{\varepsilon(i_y), y, \epsilon \sim \mathcal{N}(0,1),t[|| \epsilon - \epsilon_{\theta}(z_t, t)||^2_2]}),
\end{equation}
where $\mathcal{L}_r$ is the final reward loss function and $\mathcal{L}_{pre}$ is the loss function of SDXL. $\phi$ is a map function. $\epsilon$ is noise, $\epsilon_{\theta}$ is the denoising autoencoder and $z_t$ is the latent embedding in the denoising autoencoder. With the ReFL, we can fine-tune the SDXL to make it better align with the original and optimized prompt with PLM.
\section{Experiments}
\begin{figure*}
    \begin{subfigure}[]{0.45\textwidth}
        \centering
        \includegraphics[height=2.6in]{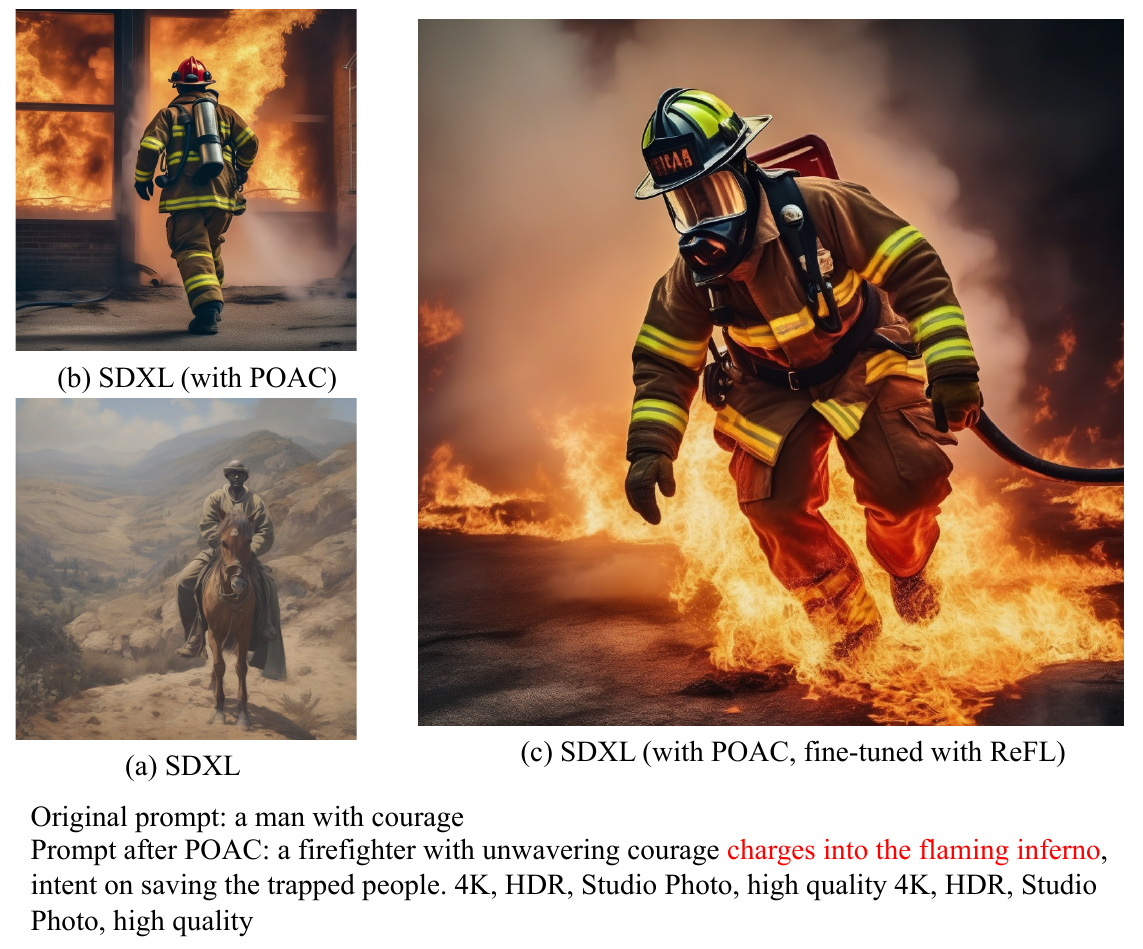}
        \caption{Courage} 
    \end{subfigure}
    \hspace{0.001\textwidth}
    \begin{subfigure}[]{0.45\textwidth}
        \centering       
        \includegraphics[height=2.6in]{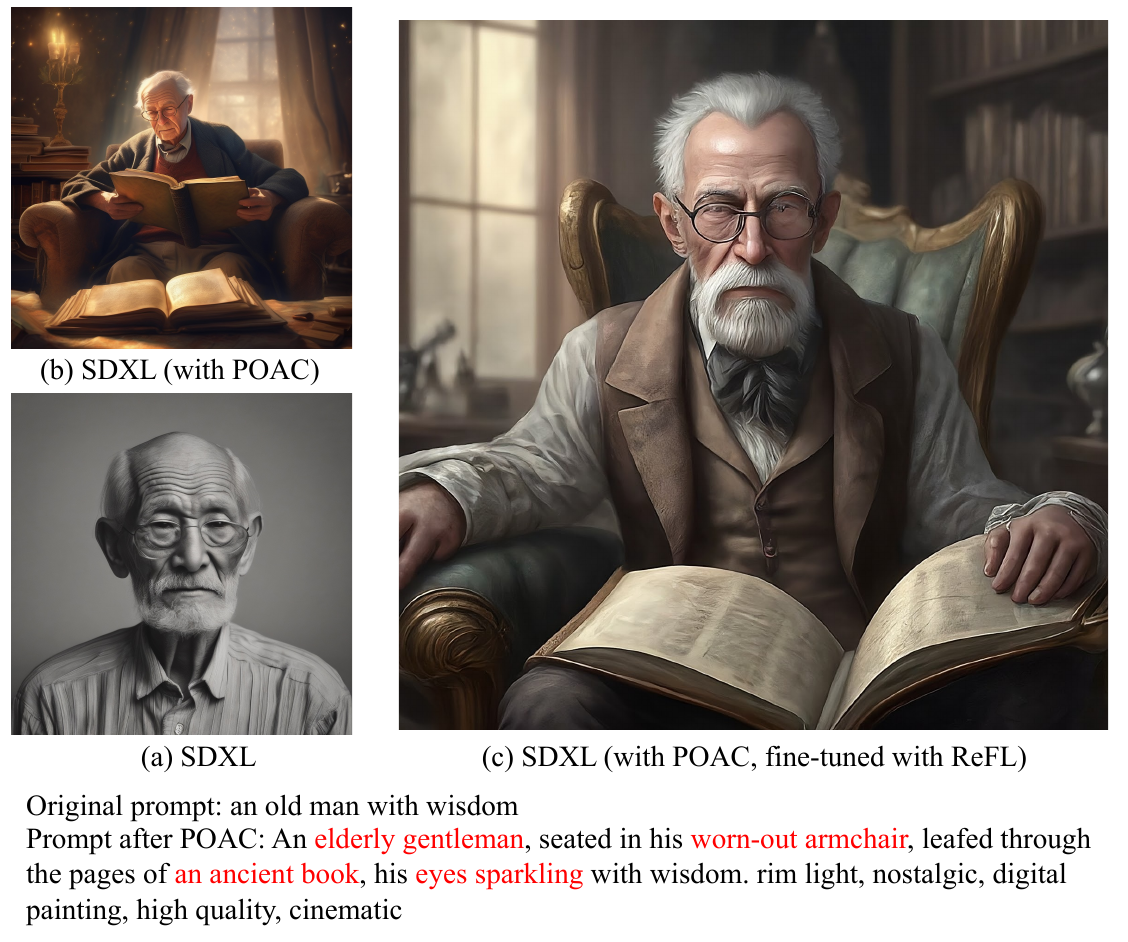}        
        \caption{Wisdom} 
    \end{subfigure}
    \hspace{0.001\textwidth}
    \begin{subfigure}[]{0.45\textwidth}
        \centering       
        \includegraphics[height=2.6in]{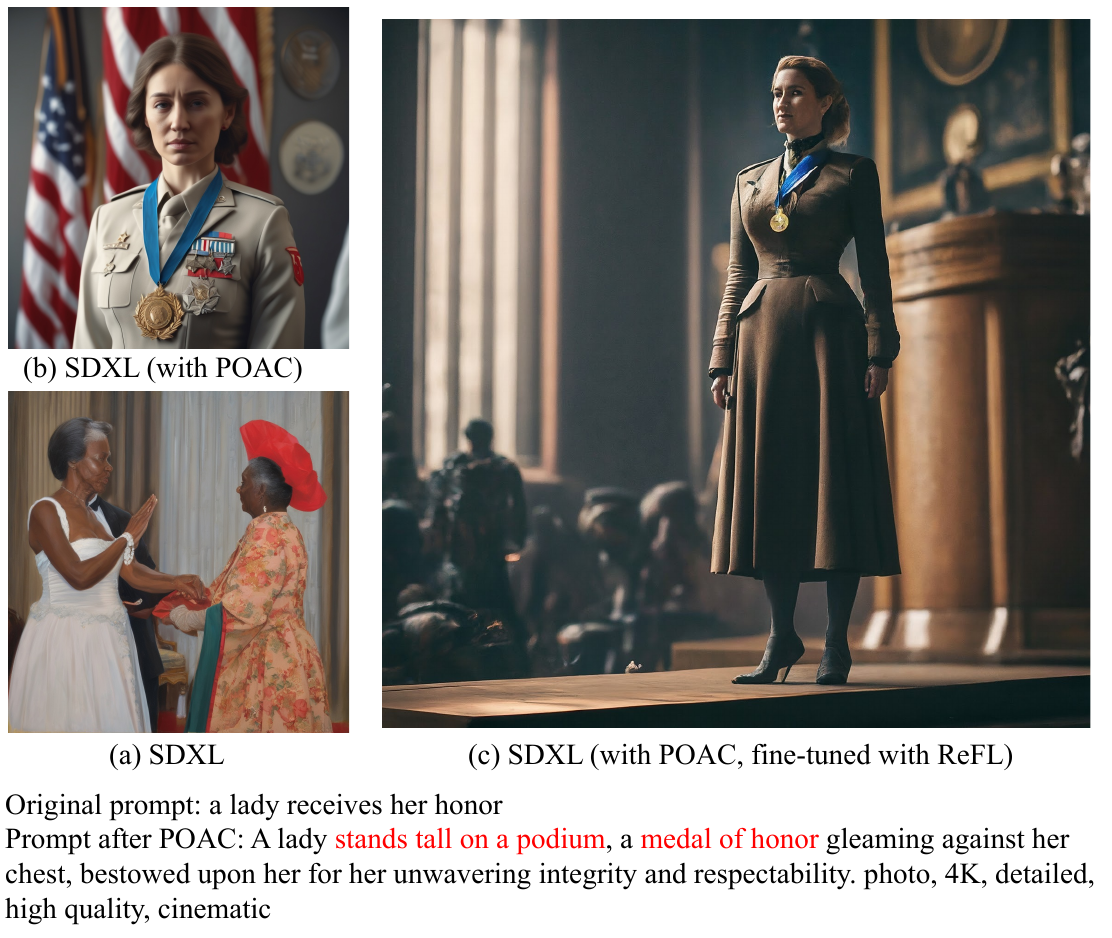}        
        \caption{Honor}     
    \end{subfigure}
    \hspace{0.001\textwidth}
    \begin{subfigure}[]{0.45\textwidth}
        \centering       
        \includegraphics[height=2.6in]{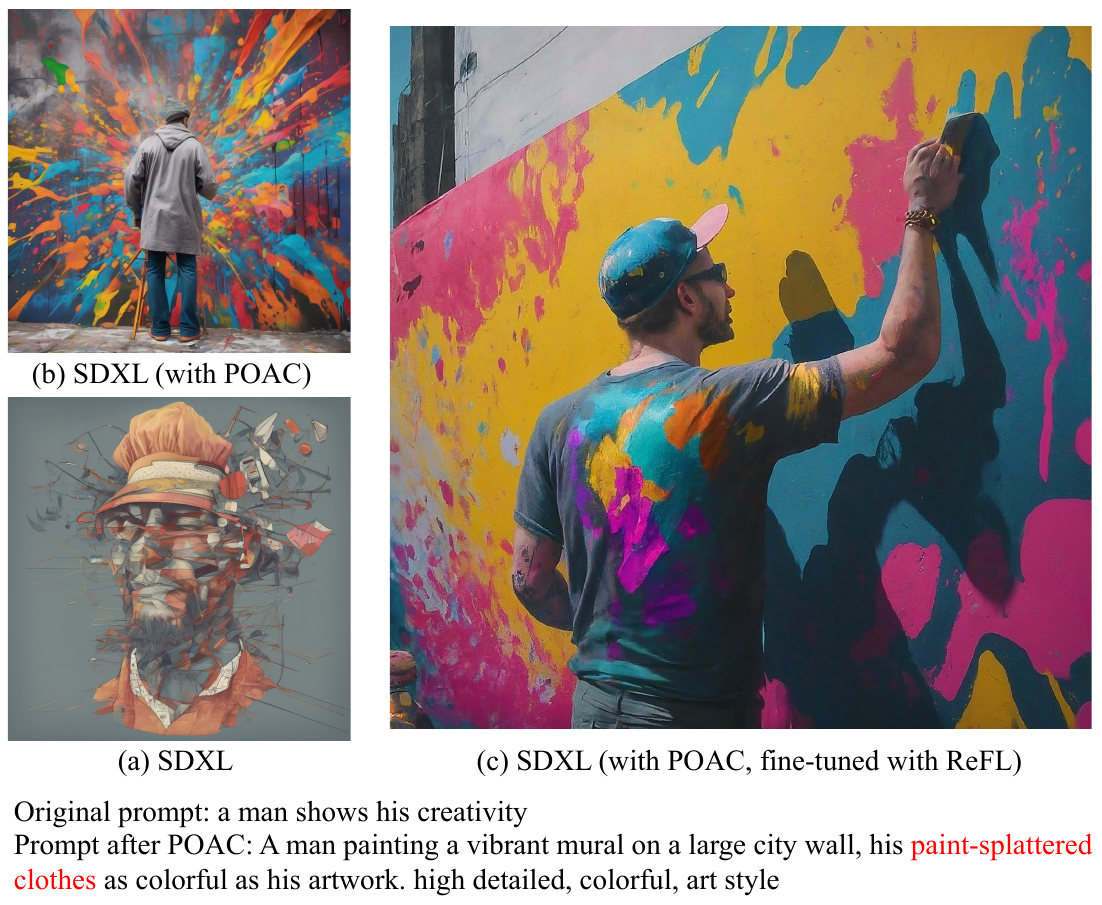}        
        \caption{Creativity}     
    \end{subfigure}
    \hspace{0.001\textwidth}
    \begin{subfigure}[]{0.45\textwidth}
        \centering       
        \includegraphics[height=2.6in]{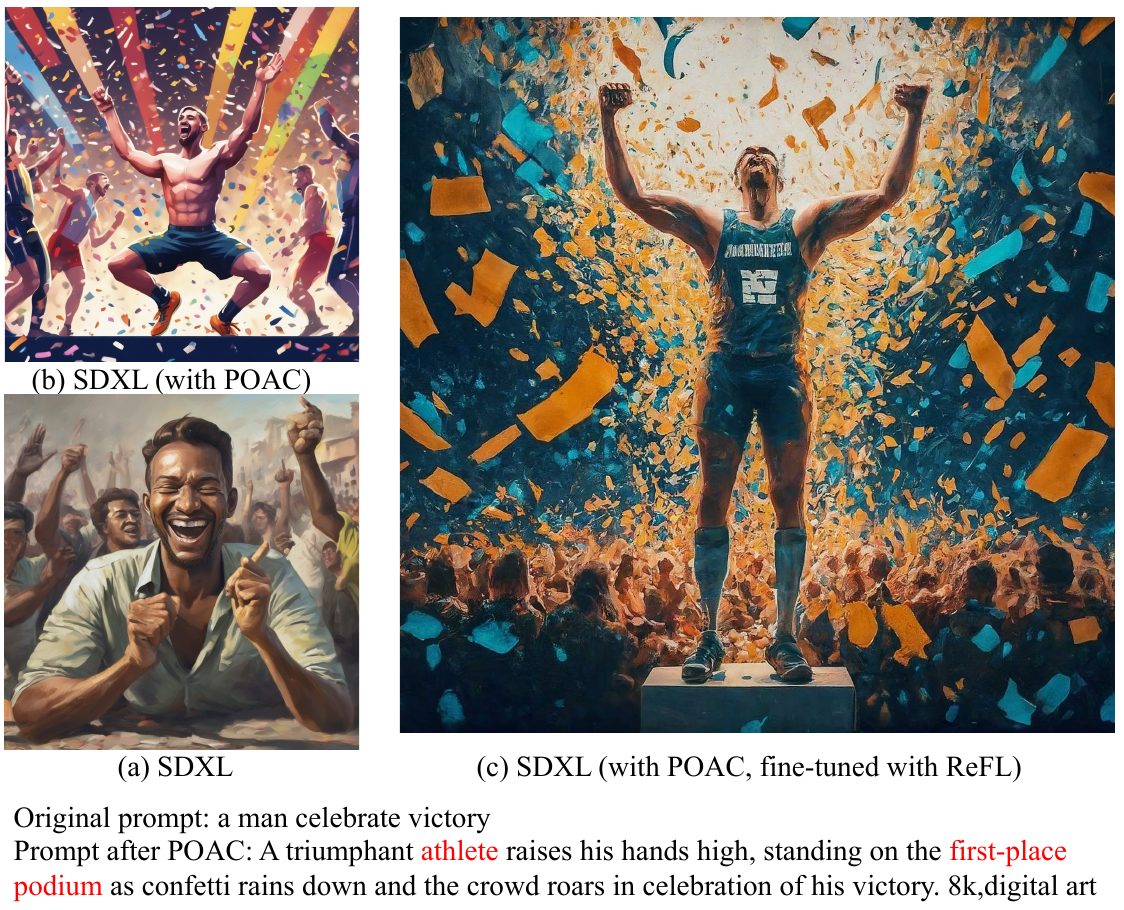}        
        \caption{Victory}     
    \end{subfigure}
    \hspace{0.001\textwidth}
    \begin{subfigure}[]{0.45\textwidth}
        \centering       
        \includegraphics[height=2.6in]{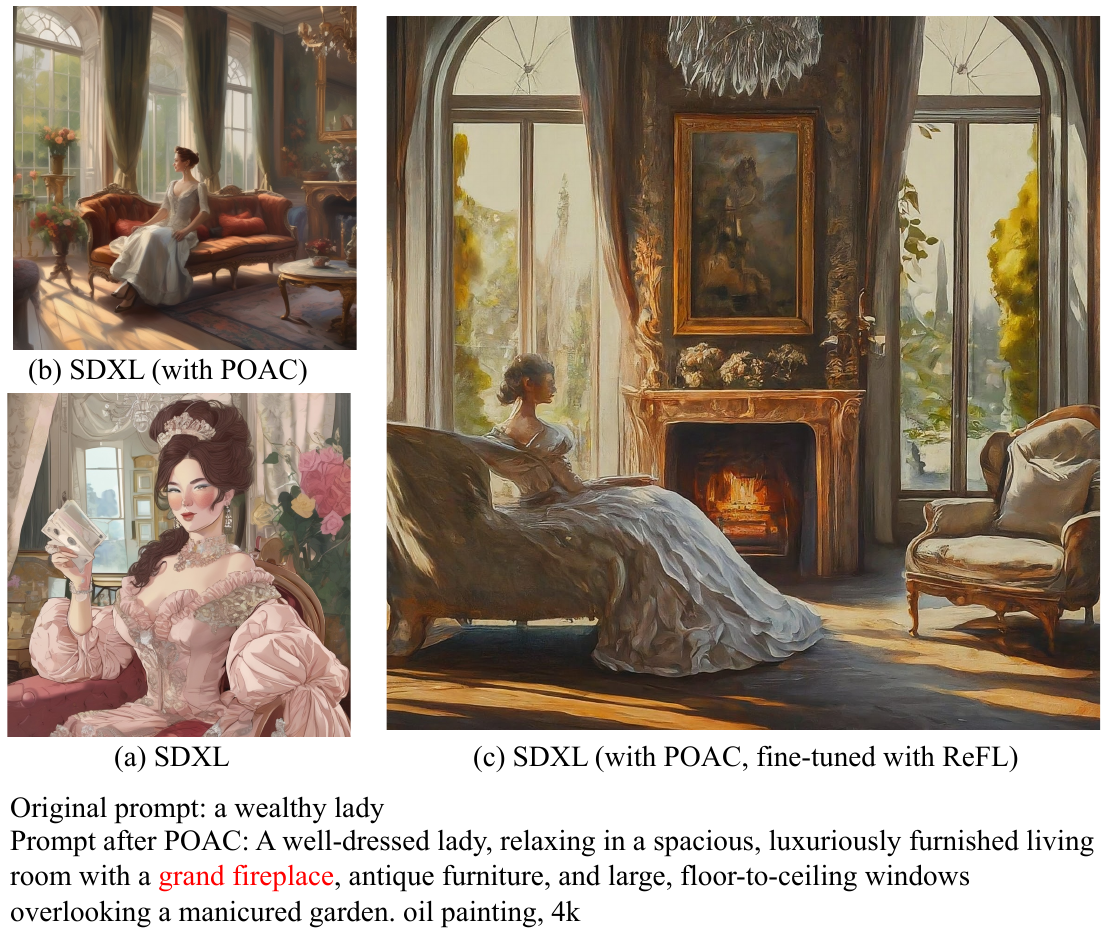}        
        \caption{Wealth}     
    \end{subfigure}
    \caption{Qualitative comparison of SDXL only, SDXL with POAC and SDXL with POAC, fine-tuned with ReFL.} 
    \label{fig:exp}
\end{figure*}

\begin{table*}[]
\begin{tabular}{l|l|l|l}
\hline
\begin{tabular}[c]{@{}l@{}}Abstract\\ concepts\end{tabular} & \multicolumn{1}{c|}{\begin{tabular}[c]{@{}c@{}}Original\\ prompts\end{tabular}} & \multicolumn{1}{c|}{Optimized prompts}                                                                                                                                                                                                        & \multicolumn{1}{c}{Modifiers}                                                                                              \\ \hline
Happiness                                                   & \begin{tabular}[c]{@{}l@{}}the essence of\\  happiness\end{tabular}             & \begin{tabular}[c]{@{}l@{}}{\color{red}A child's laughter} echoed through the {\color{red}park} as they \\ played with a {\color{red}colorful kite}, capturing the essence\\ of happiness.\end{tabular}                                                                             & \begin{tabular}[c]{@{}l@{}}Photorealistic, high-\\ detailed, 4k UHD\end{tabular}                                           \\ \hline
Anger                                                       & \begin{tabular}[c]{@{}l@{}}a man full of\\  anger\end{tabular}                  & \begin{tabular}[c]{@{}l@{}}A man, his {\color{red}face red} with anger, {\color{red}clenched his}\\  {\color{red}fists and slammed them} against the table.\end{tabular}                                                                                                             & \begin{tabular}[c]{@{}l@{}}color, depth and \\ intrigue.\end{tabular}                                                      \\ \hline
Friendship                                                  & \begin{tabular}[c]{@{}l@{}}The friendship of \\ children\end{tabular}           & \begin{tabular}[c]{@{}l@{}}Two children sat side by side on a {\color{red}wooden bench}, \\ sharing a bag of {\color{red}gummy bears} and giggling, \\ exemplifying the beauty of friendship.\end{tabular}                                                              & \begin{tabular}[c]{@{}l@{}}saturated colors, \\ high quality, nostalgic\end{tabular}                                       \\ \hline
Dream                                                       & \begin{tabular}[c]{@{}l@{}}a young man's\\ dream\end{tabular}                   & \begin{tabular}[c]{@{}l@{}}A young man's dream takes shape in the form of\\  a bustling {\color{red}startup office}, filled with {\color{red}innovative} \\ {\color{red}gadgets, whiteboards} covered in ideas, and \\ {\color{red}energetic colleagues} collaborating passionately.\end{tabular} & \begin{tabular}[c]{@{}l@{}}fujifilm xt3, outdoors, \\ beautiful lighting, raw\\  photo, 8k uhd, film  grain\end{tabular} \\ \hline

Mercy                                                  & \begin{tabular}[c]{@{}l@{}}A man shows his \\mercy. \end{tabular}           & \begin{tabular}[c]{@{}l@{}}A man, seeing a shivering {\color{red}stray dog} on the street, \\  wrapped it in his own jacket and took it to a nearby\\ {\color{red}animal shelter}.\end{tabular}   & \begin{tabular}[c]{@{}l@{}}saturated colors, \\ high quality, photorealistic\end{tabular}                                       \\ \hline
Adventure                                                & \begin{tabular}[c]{@{}l@{}}A young man is\\ undergoing an\\ adventure. \end{tabular}           & \begin{tabular}[c]{@{}l@{}}A brave young man, equipped with his {\color{red}backpack and}\\ {\color{red}torch}, is climbing a rugged {\color{red}mountain terrain} with\\ a perilous {\color{red}ravine} on one side and a dense, mysterious\\ forest on the other, under a sky ablaze with the colors\\ of sunset.\end{tabular}                                                              & \begin{tabular}[c]{@{}l@{}}saturated colors, \\ high detailed, photo,\\ 4k UHD\end{tabular}                                       \\ \hline

\end{tabular}
\caption{Prompt optimization with our proposed POAC on abstract concepts "Happiness", "Anger", "Friendship", "Dream", "Mercy" and "Adventure". The red phrases are concrete objects about the abstract concepts.}
\label{tab:prompts}
\end{table*}

We conduct experiments to validate our proposed method. Initially, we explain the implementation details in Section 4.1. Then we visualize the results and conduct comparative experiments between baseline models and our POAC framework in Section 4.2
\subsection{Implementation Detail}

\textbf{Dataset Construction}. For PLM, we collect top-500 abstract concepts from IBM Concept Abstractness\footnote{https://developer.ibm.com/exchanges/data/all/concept-abstractness/} based on their degree of abstractness. Then we manually write each concept into three different forms or a short prompt including it as the source input to PLM. Then we leverage GPT-4 to rewrite each source input into three different prompts with a dedicated scene and concrete objects correspond to their abstract concepts as target output. After experimenting with different prompt templates, we discover that the following prompt template for GPT-4 can generate the most satisfied target outputs: "Please rewrite the [Abstract Concept] in the following sentence to a
short sentence which includes a dedicated and concrete scene and also includes concrete
objects about [Source Input]".  Then we collect modifiers from user-input prompts from Lexica\footnote{https://lexica.art}. We randomly add modifiers, e.g. oil painting, nostalgic, digital painting, very intricate in Figure~\ref{fig:peace}, to the end of target prompts to ensure the diversity and quality of prompts. For reward feedback learning, the pre-training dataset is from a 625k subset of LAION-5B ~\cite{schuhmann2022laion} selected by aesthetic score.

\textbf{Supervised fine-tuning}. We fine-tune GPT-2 to predict the target prompt with the source
prompt as conditional input. The input format is [Source] Rephrase:[Target]. The model is fine-tuned with a learning rate of 5e-5, and a max length of 512 while maintaining a batch size of 64. 

\textbf{Reward Feedback Learning}. We use SDXL as our model backbone. The model is fine-tuned in half-precision with a learning rate set to 1e-5. Following The sample step range [T1, T2] is defined as [1, 10] where total steps T is set to 40. The regularizer \begin{math} \lambda \end{math} is set to 1e-3.

\subsection{Qualitative Comparison}
In Figure~\ref{fig:exp}, we illustrate the generated images of six abstract concepts: (a) courage, (b) wisdom, (c) honor, (d)creativity, (e) victory. From the images, we have the following observations.
\begin{itemize}
    \item In (a) Courage, the SDXL with the original prompt only outputs a man riding a horse, which has limited relation to courage. In SDXL (with POAC), the optimized prompt provides a scene where a firefighter faces a flaming house, which is a better expression of courage. In SDXL (with POAC, fine-tuned with ReFL), the firefighter is walking on fire, which matches the "charges into the flaming inferno" in the prompt.
    \item In (b) Wisdom, the initial output from plain SDXL only depicts an old man. However, by optimizing the prompt with POAC, the generated image now includes more specific details, such as portraying the old man reading a book, which effectively conveys the idea of wisdom. After further fine-tuning the model with ReFL, the resulting image now shows the old man dressed in a respectable suit, effectively illustrating the concept of an "elderly gentleman" as mentioned in the prompt. This enhancement ensures that the generated image better aligns with the intended representation of the prompt.
    \item In (c) Honor, SDXL generates an image of two ladies talking, which is not directly associated with honor. However, after optimizing the prompt with POAC, the revised prompt includes more specific details, such as a lady with "a medal of honor". After further fine-tuning SDXL with ReFL, the generated image now depicts a scene where "a lady stands tall on a podium", aligning better with the prompt.
    \item In (d) Creativity, the original prompt can hardly be understood by SDXL. With POAC, SDXL generates an image of a man wearing a clean jacket and painting colorful artwork. However, the prompt optimized by PLM indicates that the man's clothes are also splattered with paint. After further fine-tuning with ReFL, the model is able to generate an image of a man wearing a t-shirt that is paint-splattered.
    \item In (e) Victory, the original prompt with SDXL only generates an image of a laughing man, which is only loosely associated with victory. However, by optimizing the prompt with PLM, the revised prompt includes more concrete objects that directly depict victory, such as an "athlete" and a "first-place podium". After further fine-tuning with ReFL, the generated image now clearly shows an athlete wearing a jersey, standing on the podium, with people cheering in the background.
    \item In (f) Wealth, the image generated from the original prompt can also depict a wealthy woman holding a stack of cash. However, the overall quality of the image is poor. After optimizing the prompt, the generated image can more effectively portray the luxurious environment that she is in. The concrete object, a "grand fireplace," is now clearly visible in the image produced after fine-tuning the SDXL with ReFL.
\end{itemize}

In Table~\ref{tab:prompts}, we also provide more examples about the original prompts and optimized prompts generated by PLM about four abstract concepts: "Happiness", "Anger", "Friendship", "Dream", "Mercy" and "Adventure". The red phrases are concrete objects about the abstract concepts. From this table, we have the following findings:

(1) Our proposed POAC has the ability to transform the original prompt into a scene that utilizes concrete objects, making it easier for SDXL to comprehend. For instance, we associate "Happiness" with "a child's laughter," "Anger" with a "red face," and "Adventure" with a "backpack and torch." In our daily lives, these abstract concepts encompass human emotions and experiences, and we often rely on concrete objects to illustrate them. Through our method, we bridge the divide between human language and the abstract concepts and objects depicted in images, thereby enhancing understanding.

(2) In each optimized prompt, there are 2 or 3 concrete objects highlighted in red. These objects can be a place, an item, an expression, or an action, and they can all be effectively represented with images. By harnessing the advanced natural language understanding capabilities of the GPT-4 model, our method connects this powerful language model with a diffusion model to generate images that are more closely related to real-life scenarios.

\begin{table}[]
\begin{tabular}{l|l|l}
\hline
Models                  & \multicolumn{1}{c|}{Rel Score} & Aes Score \\ \hline
SDXL                    & 0.21                           & 0.19      \\ \hline
SDXL with POAC          & 0.29                           & 0.25      \\ \hline
SDXL with POAC and ReFL & 0.32                           & 0.26      \\ \hline
\end{tabular}

\caption{Qualitative comparison of SDXL, SDXL with POAC and SDXL with POAC and ReFL.}
\label{tab:score}
\end{table}

\subsection{Quantitative Comparison}
We also show the quantitative evaluation results of the relevance score (Rel Score) and aesthetics score (Aes Score) respectively in Table~\ref{tab:score} based on the relevance and aesthetics score in Section~\ref{sec:reward}. From this table, we can see that 
\begin{itemize}
    \item SDXL with POAC and ReFL achieves the best performance in terms of both scores, thus highlighting the efficacy of our approach. The aesthetics score of SDXL with POAC and ReFL exhibits a mere 4\% improvement when compared to SDXL with POAC alone, indicating that our method can enhance the aesthetic quality of the generated image. Furthermore, there is a noteworthy improvement of over 10\% in the relevance score, signifying that the fine-tuned SDXL model can align more effectively with the optimized prompts.
    \item SDXL with POAC, without the inclusion of ReFL, exhibits a 38\% improvement over plain SDXL. This suggests that SDXL struggles to comprehend abstract concepts, while our method greatly aids SDXL in accurately depicting the image by utilizing concrete objects to represent these abstract concepts.
\end{itemize} 
\section{Conclusion}
This study presents the development and evaluation of the Prompt Optimizer for Abstract Concepts (POAC), a pioneering method aimed at refining the capabilities of text-to-image diffusion models in accurately rendering images from abstract ideas. Leveraging a Prompt Language Model (PLM) derived from an initially pre-trained language model and further fine-tuned with a specialized dataset of abstract concept prompts created using GPT-4, our approach maps abstract concepts to detailed scenes and concrete objects. Employing a reinforcement learning-based optimization strategy, POAC aligns the stable diffusion model's image generation with the optimized prompts, significantly enhancing both the precision and the visual appeal of the output. The effectiveness of POAC is validated through qualitative and quantitative experiments, showing notable improvements in image generation tasks and proving its quality and relevance of prompts under different settings for better abstract concept visualization.

In future work, we will explore additional solutions to address the alignment problem between prompts and generated images. In addition to abstract concepts, there are other barriers that need to be overcome to produce accurate and visually pleasing images, such as biases related to race, nationality, or gender in the generated images. By improving the prompt language model's ability to optimize prompts for balanced outcomes across different groups, we can effectively address bias issues without having to train the diffusion model from scratch.

\bibliographystyle{ACM-Reference-Format}
\balance
\bibliography{sample-base}

\end{document}